\pdfoutput=1

\documentclass[11pt]{article}

\usepackage[]{ACL2023}

\usepackage{times}
\usepackage{latexsym}
\usepackage{amsmath}
\usepackage{siunitx}
\usepackage{graphicx}
\usepackage{multirow}
\usepackage{amssymb}
\usepackage{graphicx}
\usepackage{amsmath}
\usepackage{pifont}
\usepackage{caption}
\usepackage{subcaption}
\usepackage[T1]{fontenc}

\usepackage[utf8]{inputenc}

\usepackage{microtype}

\usepackage{inconsolata}

\usepackage{color}

\title{Images in Language Space: Exploring the Suitability of Large\\ Language Models for Vision \& Language Tasks}

\author{Sherzod Hakimov$^\mathbf{1}$ {\normalfont and} David Schlangen$^\mathbf{1,2}$ \\
  $^\mathbf{1}$Computational Linguistics, Department of Linguistics\\University of Potsdam, Germany \\
  $^\mathbf{2}$German Research Center for Artificial Intelligence (DFKI), Berlin, Germany \\
  \texttt{firstname.lastname@uni-potsdam.de}\\}

\begin{document}
\maketitle
\begin{abstract}

Large language models have demonstrated robust performance on various language tasks using zero-shot or few-shot learning paradigms. While being actively researched, multimodal models that can additionally handle images as input have yet to catch up in size and generality with language-only models. In this work, we ask whether language-only models can be utilised for tasks that require visual input -- but also, as we argue, often require a strong reasoning component.
Similar to some recent related work, we make visual information accessible to the language model using separate verbalisation models. Specifically, we investigate the performance of open-source, open-access language models against GPT-3 on five vision-language tasks when given textually-encoded visual information. Our results suggest that language models are effective for solving vision-language tasks even with limited samples. This approach also enhances the interpretability of a model's output by providing a means of tracing the output back through the verbalised image content.

\end{abstract}

\section{Introduction}\label{sec:introduction}

In recent years, large language models have gained significant attention in the natural language processing (NLP) community due to their impressive performance on various tasks such as machine translation, text generation, and language modelling \citep{DBLP:conf/nips/VaswaniSPUJGKP17, DBLP:conf/naacl/DevlinCLT19}. These models, which are trained on massive amounts of data, have been shown to capture complex linguistic patterns and generate coherent text~\citep{DBLP:conf/nips/BrownMRSKDNSSAA20}. Some of the most popular models are trained by OpenAI, a research organization that has released several models, including GPT~\citep{gpt}, GPT-2~\citep{gpt2}, and GPT-3~\citep{DBLP:conf/nips/BrownMRSKDNSSAA20}. In addition to GPT models, there are also many open-source or open-access large language models that researchers and organizations around the world have developed, such as BLOOM~\citep{bloom}, GPT-J~\citep{gpt-j}, OPT~\citep{opt}, Flan-T5~\citep{google-flan}.

Recent work by \citet{DBLP:journals/corr/abs-2211-09110} provided an in-depth analysis of many large language models (LLM) across 42 core scenarios.\footnote{\url{https://crfm.stanford.edu/helm/v0.2.0/}} All scenarios are language tasks that are evaluated with multiple metrics by prompting the language models with few-shots from the selected datasets, also known as \textit{in-context learning}. 
Currently, there are no comparable models directly suitable for tasks that require visual information as part of the context, 
even though such multimodal tasks have similar practical relevance.  

\begin{figure*}[ht]
  \includegraphics[width=\textwidth]{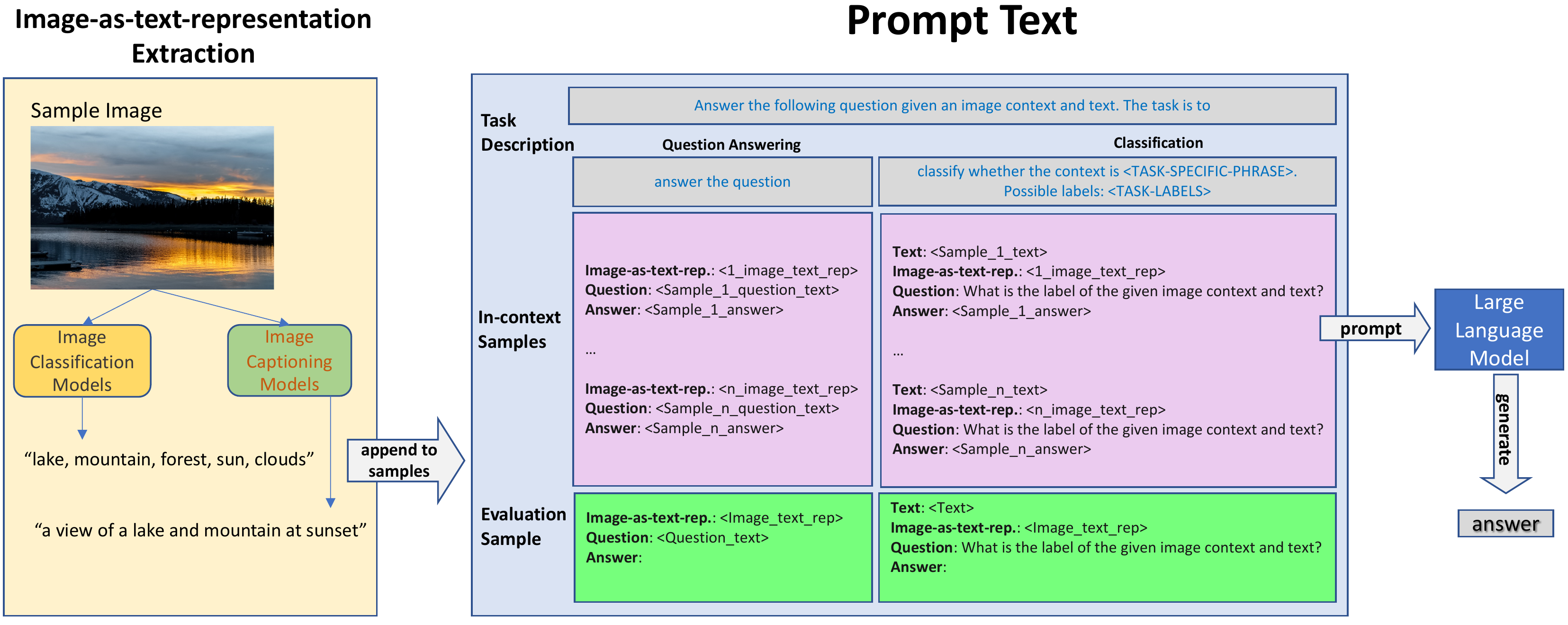}  
  \caption[width=\textwidth]{Model architecture for in-context learning for vision-language tasks. Each sample image is converted into its image-as-text-representation by applying pre-trained image captioning and classification models (yellow). The prompt text that is fed into a large language model consists of a task-specific description (blue), in-context samples (pink), and the evaluation sample (green). The language model is expected to generate a text sequence that follows the word \textit{Answer} in the evaluation sample.}
  \label{fig:model_architecture}
\end{figure*}

Pre-trained vision-language models~\citep{DBLP:journals/corr/abs-1908-03557, DBLP:conf/eccv/ChenLYK0G0020, DBLP:conf/iclr/DosovitskiyB0WZ21, DBLP:conf/icml/RadfordKHRGASAM21} have shown great promise by learning joint representation of images and text documents, but so far they have not been optimised for prompting on vision-language tasks but rather using the learned joint representations for fine-tuning on downstream tasks. 
Moreover, as we note below, many multimodal tasks appear to rely on reasoning capabilities, which larger language models have been shown to perform well on 
\citep{DBLP:journals/tacl/TalmorEGB20, li-etal-2022-pre}. Hence, in this work, we attempt to utilise such models to do in-context learning on multimodal data, achieving this by encoding the visual information in language.

While there has been some recent work going in this direction (see discussion below),
it falls short in terms of evaluating the performance of large language models across multiple dimensions, applying them to a diverse range of vision-language tasks, and comparing the performance of GPT models with open-source or open-access models. As such, there is a need for further research in this area to fully understand the capabilities and limitations of these models for vision-language tasks.

In this paper, we aim to explore the capabilities of large language models for in-context learning and their potential to improve performance on multimodal tasks (see Figure~\ref{fig:model_architecture}). To this end, we conducted a series of experiments to evaluate the performance of various language models (closed, open-source and open-access) on tasks that involve multiple modalities, such as vision and language. The tasks vary from identification of hate speech and sentiment to visual reasoning and question answering. Our results provide insights into the strengths and limitations of these models and highlight methods of expressing visual content through textual descriptions. Our work aims to analyse how much general reasoning is in language models by evaluating them on multimodal tasks where the visual content is only accessible partially or indirectly (since the visual content is verbalised and represented in textual form) accessible. Our main contributions %
are as follows\footnote{Source code and all resources are made publicly available at \url{https://github.com/clp-research/language-models-multimodal-tasks}}:

\begin{itemize}
\itemsep0em 
    \item We examine the impact of in-context learning with large language models on five vision-language tasks, four classification and one question answering tasks;
    \item We investigate the impact of the textual description generation for the visual content on the model performance for the respective tasks;
    \item We compare the performance of open-source and open-access models with GPT-3 on the selected vision-language tasks.
\end{itemize}

\section{Related Work}\label{sec:related_work}

Recent work by \citet{DBLP:journals/corr/abs-2211-09110} provided an in-depth analysis of many 34 large language models (LLM), open, limited-access, and closed. Their analysis revealed the capabilities and limitations of these models across 42 core scenarios. All scenarios are language tasks that are evaluated with 57 metrics by prompting the language models with a few shots from the selected datasets. Such a way of leveraging pre-trained language models for downstream tasks is known as \textit{in-context learning}, where a certain task description and a few shots are presented as a context for the model. A recent survey by \citet{dong2022survey} describes the developed techniques for in-context learning where they present a taxonomy that divides the techniques used for prompting such as selection of in-context samples, reasoning step by step (chain of thought)~\citep{DBLP:journals/corr/abs-2201-11903}, task definition, etc. Moreover, \citet{DBLP:journals/corr/abs-2202-12837} assessed the importance of choosing the in-context samples and its effect on the performance.

So far, a large-scale analysis of large language models and their performances for multimodal data, such as vision-language tasks, has not been done. A handful of methods demonstrated the effectiveness of in-context learning for multimodal tasks. \citet{DBLP:journals/ijcv/ZhouYLL22, DBLP:conf/cvpr/ZhouYL022}  modelled the context words in prompts for applying pre-trained vision-language tasks for downstream vision tasks. \citet{DBLP:conf/nips/TsimpoukelliMCE21} trained a vision encoder to represent images as a sequence of continuous embeddings where a prompted pre-trained language model generates a caption. \citet{DBLP:conf/aaai/YangGW0L0W22} demonstrated the applicability of GPT-3 on a visual question answering task where they converted the images into textual descriptions by using an image captioning model and extraction of visual tags that correspond to detected objects, landmarks, person, image type, etc. \citet{DBLP:journals/corr/abs-2204-00598} follows similar methodology by showing applications on multiple applications that include modalities such as audio, video beside image and text. \citet{gui-etal-2022-kat}'s method is complementary to the previous method with an addition of a contrastive learning module that retrieves knowledge entries from Wikidata knowledge graph~\citep{DBLP:journals/cacm/VrandecicK14}. \citet{DBLP:journals/corr/abs-2205-10747} applied the method of converting images into textual descriptions to video tasks. The resulting outputs are temporally aligned for a video and then fed into GPT-3 with few shots. More recently, \citet{merullo2023linearly} aligned image-text encoders by training a linear project layer and keeping the pre-trained image and text encoders frozen. Our paper presents a study that goes beyond these similar approaches by extending the experimental evaluation to multiple datasets, comparing open-source language models with GPT-3, and evaluating different methods of acquiring textual representation for the visual content.

\section{Text-Visual In-Context Learning}\label{sec:methodology}

In this section, we describe the proposed methodology of applying \textit{in-context learning} to vision-language tasks. In-context learning essentially works by prompting a pre-trained language model with the task and expecting it to generate text that solves a particular task. It is performed by giving a few-shots of the respective task at inference time without requiring updating the model weights and expecting the model to generate text corresponding to the expected output.

Formally, given a query input text $\mathbf{x}$ and a set of candidate answers $Y =\{\mathbf{y_1} ... \mathbf{y_m}\}$, which can be class labels for a particular task or free text, a pre-trained model \textit{M} outputs a candidate answer with the maximum score conditioned on the task description $T$, \textit{n} in-context sample pairs $C = \{  (\mathbf{x}_1, \mathbf{y}_1) ... (\mathbf{x}_n, \mathbf{y}_n) \}$. The likelihood of the candidate answer $\mathbf{y_j}$ can be represented by a scoring function $f$ with the language model \textit{M}~\citep{DBLP:conf/iclr/WeiBZGYLDDL22, dong2022survey}:

\begin{equation}
    \text{P} (\mathbf{y_j}|\mathbf{x}) \triangleq f_M (\mathbf{y_j}, T, C, \mathbf{x})
\end{equation}

The final predicted candidate answer of the model ($\hat{\mathbf{y}}$) can be formulated as:

\begin{equation}
    \hat{\mathbf{y}} = \text{argmax}_{\mathbf{y_j} \in Y} \text{P}(\mathbf{y_j}|\mathbf{x})
\end{equation}

Our proposed methodology for ``text-visual in-context learning'' is shown in Figure~\ref{fig:model_architecture}. First, all images from all evaluated datasets have been passed through multiple pre-trained image models to obtain the textual description of the visual content, which we refer to as \textit{image-as-text-representation} throughout the paper. The image-as-text-representation is essentially a textual description of the visual content that captures important visual aspects. The prompt text comprises the task description, in-context sample pairs, and the input of the evaluation sample. Given such a prompt text, the language model generates a sequence of text tokens as an output.

We evaluate the proposed methodology on various vision-language datasets that include either classification or question answering tasks. Thus, the task description is different between these two categories. The task description for the classification tasks is further replaced with the task-specific phrase that describes the downstream task and provides the task-specific class labels. More details on the exact prompt text for each dataset are provided in Appendix~\ref{sec:prompt_appendix}. Next, we describe the methods for extracting image-as-text-representation, selecting in-context samples, and aggregating answer predictions in cases where the language model is prompted multiple times with various in-context samples for the evaluation sample.

\subsection{Image-as-Text-Representation Extraction}
We use two different methods to extract textual representation of images for any vision-language task. The first  is to use pre-trained image captioning models that generate a text sequence describing the input image. The second  is to employ multiple pre-trained image classification models and extract top-scoring class labels. The extracted class labels from all models are merged to form the set of \textit{visual tags} that describe the image. Specifically, we use pre-trained models to recognise objects, indoor or outdoor scenes, and persons and their facial emotions. These methods yield a different textual description of an input image, which is used as \textit{image-as-text-representation} in the prompt text.

\subsection{In-Context Sample Selection}\label{subsec:sample_selection}

The selection of samples for in-context learning is an essential step for prompting large language models. Each model has its own limitation of the maximum input tokens that a prompt text can have (e.g. 512 tokens for Flan-T5, 4000 for GPT-3). Therefore, only a few samples can be used (few-shot), and the selection directly impacts the model performance~\citep{DBLP:conf/aaai/YangGW0L0W22, DBLP:journals/corr/abs-2202-12837}. We experiment with the following sample selection methods.

\noindent \textbf{Random sample selection} works by selecting any random \textit{n} samples from the training split of a dataset (that fit into the maximum input token length of a language model). 

\noindent \textbf{Adaptive sample selection} uses specific similarity measurement to rank samples with respect to the input sample. Top-ranking \textit{n} samples are selected (that fit into the maximum input token length of a language model) to prompt a language model.

The in-context samples are selected from the training split of the respective dataset.

\section{Experimental Setup}\label{sec:experimental_setup}

In this section, we describe the details of the building blocks of the methodology, the evaluated datasets, large language models, and methods for obtaining a textual description of images.

\subsection{Datasets, Comparison Models \& Evaluation Metrics}\label{subsec:datasets}

We use the following five datasets to evaluate the performance of the closed and open-access language models. The best-performing prior models are selected from the leaderboards of the respective datasets. These models are used for comparison with our prompting method. 

\begin{itemize}
\itemsep0em 
    \item \textit{MAMI - Multimedia Automatic Misogyny Identification} (SemEval 2022 Task 5)~\citep{mami}: the dataset consists of \textit{memes} that are classified for being offensive hateful towards women. The train and test splits have \num{10000} and \num{1000} samples, respectively. We use the sub-task A for the experiments to predict a binary class indicating whether the given meme is misogynous. \\
    \textbf{Comparison model}: \citet{zhang-wang-2022-srcb} proposed to use an ensemble system composed of pre-trained models (CLIP) used for extracting features from multimodal data.\\ 
    \textbf{Evaluation metric}: Macro-F1
    
    \item \textit{HF - Hateful Memes}~\citep{DBLP:conf/nips/KielaFMGSRT20}: is another dataset that focuses on classifying \textit{memes} whether the overall message it is hateful or not towards any group. We use the provided development split for the evaluation since the test split is closed to the community at the time of writing. The train and development splits have \num{8500} and \num{500} samples, respectively.
    \\
    \textbf{Comparison model}: the best-performing model provided is by \citet{DBLP:conf/nips/KielaFMGSRT20}, for which the model performance on the development split is available. The method uses  pre-trained ViLBERT model that is later fine-tuned on the dataset.\\ 
    \textbf{Evaluation metric}: Accuracy

    \item \textit{MVSA - Multi-View Sentiment Analysis}~\citep{DBLP:conf/mmm/NiuZPE16}: is a multimodal sentiment analysis dataset collected from Twitter. 
    The task is to classify the sentiment of the given post with an image and tweet text into \textit{positive}, \textit{negative}, or \textit{neutral}. 
    
    Previous work on this dataset has used different train and test splits, making the direct comparison among approaches not feasible. We follow recently provided splits by \citet{DBLP:conf/mir/CheemaHME21} and their evaluation scheme by performing 10-fold cross-validation on the respective train and test splits. Overall, the dataset includes total \num{3928} samples with 
\num{2328}, \num{1167}, \num{433} samples corresponding to \textit{positive}, \textit{negative}, and \textit{neutral} class labels, respectively. We use the version named \textit{MVSA-Single} of this dataset.
    \\
    \textbf{Comparison model}: \citet{DBLP:conf/mir/CheemaHME21}'s model uses image features from CLIP and text features from RoBERTa models and fine-tune them on the dataset. \\ 
    \textbf{Evaluation metric}: Accuracy averaged over 10-folds.
    
    \item \textit{OK-VQA - Outside Knowledge Visual Question Answering}~\citep{DBLP:conf/cvpr/MarinoRFM19}:  is a visual question answering dataset consisting of \num{14055} samples where each sample contains an open-ended question and five ground truth answers. The task is to predict one of the expected answers given the question and the image. There are \num{9009} and \num{5046} samples in the train and test splits of the dataset, respectively.    \\
    \textbf{Comparison model}: \citet{DBLP:conf/aaai/WuLSM22}'s method is based on  three-stage scheme where the first step generates a set of answer candidates by analysing the syntactic structure of the question. The next step retrieves candidate answers by searching the Wikipedia and ConceptNet, and finally the third step validates the candidate answers.\\ 
    \textbf{Evaluation metric}: Accuracy

    \item \textit{NLVR2 - Natural Language for Visual Reasoning for Real}~\citep{DBLP:conf/acl/SuhrZZZBA19}: is a dataset for reasoning over two images and a statement where the task is predict whether the statement is \textit{true} or \textit{false}. The dataset includes \num{86373} and \num{6967} samples for the train and test splits, respectively. We used the \textit{test-public} split of the dataset. \\
    \textbf{Comparison model}: \citet{DBLP:conf/eccv/ChenLYK0G0020}'s approach is based on first pre-training a joint multimodal model on image captioning datasets and then fine-tune the model on the task.\\ 
    \textbf{Evaluation metric}: Accuracy

\end{itemize}

\subsection{Language Models}

We experiment with multiple pre-trained open-source and open-access language models and compare them against GPT-3. These language models are as follows:

\begin{itemize}
\itemsep-0.2em 
    \item \textit{Flan-T5}~\citep{google-flan}: is a language model fine-tuned on multiple tasks with an instruction-specific training paradigm. We use the \textit{flan-t5-xxl} version.%

    \item \textit{T0pp}~\citep{DBLP:conf/iclr/SanhWRBSACSRDBX22}: is a language model that has been fine-tuned on multiple datasets to perform for zero or few-shot prompting.

    \item \textit{OPT}~\citep{opt}: is a language model trained on multiple large datasets. The language model has various versions with different sizes. We use the \textit{opt-2.7b} version.%

    \item \textit{GPT-3}: we use the \textit{text-davinci-003} version.
    
\end{itemize}

\subsection{Methods for Extracting Image-as-Text-Representations}\label{subsec:image_context_methods}

The generation of the textual representation of images is carried out in two ways: image captioning and the combination of multiple image classification model outputs.

\textbf{Image Captioning}: we use the following image captioning models to convert the images to textual descriptions: 

\begin{itemize}
\itemsep-0.2em 
    \item ViT-GPT-2 (Vision Transformers GPT-2)~\citep{nlp_connect_2022}%
    \item OFA (One for all)~\citep{DBLP:conf/icml/WangYMLBLMZZY22}%
    \item BLIP (Bootstrapping Language-Image Pre-training)~\citep{blip}%
\end{itemize}

\textbf{Visual Tags}: we use the following image classification models to build the set of tags extracted from a given image:

\begin{itemize}
\itemsep-0.3em 
    \item \textit{Image type}: a zero-shot classification with CLIP~\citep{DBLP:conf/icml/RadfordKHRGASAM21} by pairing an image with one of the following text snippets and selecting the one that outputs the highest probability: ``This is an image``, ``This is a sketch``, ``This is a cartoon``, ``This is a painting``. We select the top-ranking class label that has a probability higher or equal to \num{0.80}.
    
    \item \textit{Object}: the pre-trained Detection Transformer (DETR) model~\citep{DBLP:conf/eccv/CarionMSUKZ20} is used to obtain the bounding boxes of detected objects. We select the top-ranking class labels that have a probability higher or equal to \num{0.90}.

    \item \textit{Indoor and outdoor scenes}: we use two different pre-trained models to predict the scenes in the given images. The first model is Vision Transformer (ViT)~\citep{DBLP:journals/corr/abs-2006-03677} pre-trained on Indoor Scene dataset~\citep{DBLP:conf/cvpr/QuattoniT09}. The second model is a pre-trained ResNet-50 on Places365 dataset~\citep{DBLP:journals/pami/ZhouLKO018}. We select the top-ranking class labels that have a probability higher or equal to \num{0.80}.

    \item \textit{Facial expression}: we use the pre-trained MTCNN model~\citep{7553523} to detect faces in images and identify seven facial emotions: angry, disgust, fear, happy, sad, surprise, neutral.~\citep{goodfellow2015challenges}. We select the top-ranking detected faces (probability >= \num{0.90}) and use them to infer the facial expression classes. The top-ranking facial expression class label (probability >= \num{0.50}) is selected for each detected face.
    
\end{itemize}

\subsection{Prompt Structure}

\textbf{Similarity measurement}: As mentioned above in Section~\ref{subsec:sample_selection}, we employ two different methods for selecting samples for in-context learning: random and adaptive. In order to select the best fitting \textit{n} samples for the adaptive prompting, we use the Sentence Transformers~\citep{DBLP:conf/emnlp/ReimersG19} to calculate the similarities among samples for the adaptive method. The pre-trained \textit{all-mpnet-base-v2} model is used to extract embeddings from two given sample documents and calculates the cosine similarity between them\footnote{\url{https://huggingface.co/sentence-transformers/all-mpnet-base-v2}}. For any given two samples (one evaluation and the other one from a training split), we calculate the similarity between the text content and image-as-text representation obtained from the methods described before. The similarities from textual content and image-as-text-representation are averaged.

\textbf{Sample selection}: Once the most similar samples to the given evaluation sample are identified, the next step is to select \textit{n} samples out of them. During selection, we ensure that the selected samples are equally distributed across the class labels for any dataset. This only applies to the classification tasks where the labels are predefined (e.g. hateful or not, true/false, positive/negative/neutral). It is to present samples with different labels for in-context learning. We also experiment with a zero-shot (\textit{n}=0) where the prompted text includes only the task description.

The prompt structure for each dataset is available in Appendix~\ref{sec:prompt_appendix}.

\subsection{Model Parameters \& Implementation}

We experimented with various configurations of the model parameters. The following values are used for all language models: \textit{max new tokens} set as \num{10}, \textit{number of beams} is set as \num{10}, \textit{temperature} is set to the default value of each language model.

The implementation of the overall architecture and other building blocks (image captioning \& classification) is based on the PyTorch library. We used the language models that are available in the HuggingFace directory and queried the backend API of OpenAI for experiments with GPT-3. All experiments have been carried out on two NVIDIA A100 GPUs (80 GB). The estimated runtime of all experiments is approximately 200 hours.

\begin{table}[!ht]
\centering
\small
\begin{tabular}{|ll|cccc|c|}
\hline
\multicolumn{2}{|l|}{}                                          & \multicolumn{4}{c|}{\textbf{Number of Samples}}                                                                                   &                                                                     \\ \hline
\multicolumn{1}{|l|}{\textbf{Dataset}}        & \textbf{Models} & \multicolumn{1}{c|}{\textbf{n=0}} & \multicolumn{1}{c|}{\textbf{n=1}} & \multicolumn{1}{c|}{\textbf{n=2}} & \textbf{n=3}          & \textbf{S} \\ \hline

\multicolumn{1}{|l|}{\multirow{6}{*}{\begin{tabular}[c]{@{}c@{}}\textbf{MVSA} \\ (Acc)\end{tabular}}}   & \textbf{Flan-T5} & \multicolumn{1}{c|}{6.5}  & \multicolumn{1}{c|}{22.9}  & \multicolumn{1}{c|}{28.4} & 29.1  & \textit{r} \\ \cline{2-7} 
\multicolumn{1}{|l|}{}   & \textbf{Flan-T5} & \multicolumn{1}{c|}{6.5}  & \multicolumn{1}{c|}{21.4}  & \multicolumn{1}{c|}{31.8} & 35.2  & \textit{a} \\ \cline{2-7} 
\multicolumn{1}{|l|}{}   & \textbf{T0pp} & \multicolumn{1}{c|}{\textbf{68.1}}  & \multicolumn{1}{c|}{54.2}  & \multicolumn{1}{c|}{57.7} & 45.0  & \textit{r} \\ \cline{2-7} 
\multicolumn{1}{|l|}{}   & \textbf{T0pp} & \multicolumn{1}{c|}{68.1}  & \multicolumn{1}{c|}{57.4}  & \multicolumn{1}{c|}{62.3} & 51.8  & \textit{a} \\ \cline{2-7} 
\multicolumn{1}{|l|}{}   & \textbf{OPT} & \multicolumn{1}{c|}{0.0}  & \multicolumn{1}{c|}{12.9}  & \multicolumn{1}{c|}{11.8} & 14.4  & \textit{r} \\ \cline{2-7} 
\multicolumn{1}{|l|}{}   & \textbf{OPT} & \multicolumn{1}{c|}{0.0}  & \multicolumn{1}{c|}{11.5}  & \multicolumn{1}{c|}{11.1} & 11.3  & \textit{a} \\ \hline

\multicolumn{1}{|l|}{\multirow{6}{*}{\begin{tabular}[c]{@{}c@{}}\textbf{OK-VQA} \\ (Acc)\end{tabular}}}   & \textbf{Flan-T5} & \multicolumn{1}{c|}{27.0}  & \multicolumn{1}{c|}{28.1}  & \multicolumn{1}{c|}{28.7} & 29.5  & \textit{r} \\ \cline{2-7} 
\multicolumn{1}{|l|}{}   & \textbf{Flan-T5} & \multicolumn{1}{c|}{27.0}  & \multicolumn{1}{c|}{32.4}  & \multicolumn{1}{c|}{34.4} & \textbf{35.4}  & \textit{a} \\ \cline{2-7} 
\multicolumn{1}{|l|}{}   & \textbf{T0pp} & \multicolumn{1}{c|}{13.9}  & \multicolumn{1}{c|}{18.4}  & \multicolumn{1}{c|}{18.4} & 18.2  & \textit{r} \\ \cline{2-7} 
\multicolumn{1}{|l|}{}   & \textbf{T0pp} & \multicolumn{1}{c|}{13.9}  & \multicolumn{1}{c|}{19.4}  & \multicolumn{1}{c|}{20.2} & 20.3  & \textit{a} \\ \cline{2-7} 
\multicolumn{1}{|l|}{}   & \textbf{OPT} & \multicolumn{1}{c|}{0.9}  & \multicolumn{1}{c|}{5.7}  & \multicolumn{1}{c|}{3.6} & 4.3  & \textit{r} \\ \cline{2-7} 
\multicolumn{1}{|l|}{}   & \textbf{OPT} & \multicolumn{1}{c|}{0.9}  & \multicolumn{1}{c|}{10.0}  & \multicolumn{1}{c|}{3.9} & 3.3  & \textit{a} \\ \hline

\multicolumn{1}{|l|}{\multirow{6}{*}{\begin{tabular}[c]{@{}c@{}}\textbf{NLVR2} \\ (Acc)\end{tabular}}}   & \textbf{Flan-T5} & \multicolumn{1}{c|}{0.0}  & \multicolumn{1}{c|}{12.0}  & \multicolumn{1}{c|}{19.5} & 19.7  & \textit{r} \\ \cline{2-7} 
\multicolumn{1}{|l|}{}   & \textbf{Flan-T5} & \multicolumn{1}{c|}{0.0}  & \multicolumn{1}{c|}{25.6}  & \multicolumn{1}{c|}{31.7} & 31.7  & \textit{a} \\ \cline{2-7} 
\multicolumn{1}{|l|}{}   & \textbf{T0pp} & \multicolumn{1}{c|}{58.6}  & \multicolumn{1}{c|}{57.5}  & \multicolumn{1}{c|}{49.2} & 49.5  & \textit{r} \\ \cline{2-7} 
\multicolumn{1}{|l|}{}   & \textbf{T0pp} & \multicolumn{1}{c|}{\textbf{58.6}}  & \multicolumn{1}{c|}{55.2}  & \multicolumn{1}{c|}{50.7} & 50.1  & \textit{a} \\ \cline{2-7} 
\multicolumn{1}{|l|}{}   & \textbf{OPT} & \multicolumn{1}{c|}{42.2}  & \multicolumn{1}{c|}{47.7}  & \multicolumn{1}{c|}{46.3} & 45.3  & \textit{r} \\ \cline{2-7} 
\multicolumn{1}{|l|}{}   & \textbf{OPT} & \multicolumn{1}{c|}{42.2}  & \multicolumn{1}{c|}{26.9}  & \multicolumn{1}{c|}{10.8} & 8.1  & \textit{a} \\ \hline

\multicolumn{1}{|l|}{\multirow{6}{*}{\begin{tabular}[c]{@{}c@{}}\textbf{HF} \\ (Acc)\end{tabular}}}   & \textbf{Flan-T5} & \multicolumn{1}{c|}{55.2}  & \multicolumn{1}{c|}{53.2}  & \multicolumn{1}{c|}{53.6} & 53.8  & \textit{r} \\ \cline{2-7} 
\multicolumn{1}{|l|}{}   & \textbf{Flan-T5} & \multicolumn{1}{c|}{55.2}  & \multicolumn{1}{c|}{\textbf{57.4}}  & \multicolumn{1}{c|}{56.6} & 56.0  & \textit{a} \\ \cline{2-7} 
\multicolumn{1}{|l|}{}   & \textbf{T0pp} & \multicolumn{1}{c|}{50.0}  & \multicolumn{1}{c|}{48.8}  & \multicolumn{1}{c|}{1.6} & 0.0  & \textit{r} \\ \cline{2-7} 
\multicolumn{1}{|l|}{}   & \textbf{T0pp} & \multicolumn{1}{c|}{50.0}  & \multicolumn{1}{c|}{53.6}  & \multicolumn{1}{c|}{49.0} & 33.0  & \textit{a} \\ \cline{2-7} 
\multicolumn{1}{|l|}{}   & \textbf{OPT} & \multicolumn{1}{c|}{0.2}  & \multicolumn{1}{c|}{44.4}  & \multicolumn{1}{c|}{41.4} & 36.0  & \textit{r} \\ \cline{2-7} 
\multicolumn{1}{|l|}{}   & \textbf{OPT} & \multicolumn{1}{c|}{0.2}  & \multicolumn{1}{c|}{35.2}  & \multicolumn{1}{c|}{38.2} & 30.6  & \textit{a} \\ \hline

\multicolumn{1}{|l|}{\multirow{6}{*}{\begin{tabular}[c]{@{}c@{}}\textbf{MAMI} \\ (F1)\end{tabular}}}   & \textbf{Flan-T5} & \multicolumn{1}{c|}{41.5}  & \multicolumn{1}{c|}{51.7}  & \multicolumn{1}{c|}{37.7} & 56.1  & \textit{r} \\ \cline{2-7} 
\multicolumn{1}{|l|}{}   & \textbf{Flan-T5} & \multicolumn{1}{c|}{41.5}  & \multicolumn{1}{c|}{61.9}  & \multicolumn{1}{c|}{\textbf{64.4}} & 64.1  & \textit{a} \\ \cline{2-7} 
\multicolumn{1}{|l|}{}   & \textbf{T0pp} & \multicolumn{1}{c|}{26.0}  & \multicolumn{1}{c|}{22.2}  & \multicolumn{1}{c|}{6.8} & 0.0  & \textit{r} \\ \cline{2-7} 
\multicolumn{1}{|l|}{}   & \textbf{T0pp} & \multicolumn{1}{c|}{26.0}  & \multicolumn{1}{c|}{46.0}  & \multicolumn{1}{c|}{33.8} & 21.8  & \textit{a} \\ \cline{2-7} 
\multicolumn{1}{|l|}{}   & \textbf{OPT} & \multicolumn{1}{c|}{17.0}  & \multicolumn{1}{c|}{22.2}  & \multicolumn{1}{c|}{22.1} & 21.9  & \textit{r} \\ \cline{2-7} 
\multicolumn{1}{|l|}{}   & \textbf{OPT} & \multicolumn{1}{c|}{17.0}  & \multicolumn{1}{c|}{22.3}  & \multicolumn{1}{c|}{22.0} & 22.4  & \textit{a} \\ \hline

\end{tabular}
\caption{Ablation study on the number of in-context samples and the method of selecting them. \textit{n} refers to the number of samples in a given prompt, \textbf{S} stands for the selection method: adaptive (\textit{a}) or random (\textit{r}). All runs are based on using captioning from BLIP model. The best result for each dataset is highlighted in bold.}
\label{tab:in_context_samples}
\end{table}

\section{Results and Analysis}\label{sec:results}

In this section, we discuss the obtained results from the experiments such as the impact of in-context sample selection, image-as-text representation methods and comparing with fine-tuned vision-language models on the selected datasets.

\begin{table}[!ht]
\small
\begin{tabular}{|l|l|c|c|c|c|}
\hline

\multicolumn{1}{|c|}{\multirow{2}{*}{\textbf{Dataset}}} & \multicolumn{1}{l|}{\multirow{2}{*}{\textbf{Models}}} & \multicolumn{4}{c|}{\textbf{\begin{tabular}[c]{@{}c@{}}Image-as-text\\ Representation\end{tabular}}}                                          \\ \cline{3-6} 
\multicolumn{1}{|c|}{}                                  & \multicolumn{1}{l|}{}                                 & \multicolumn{1}{c|}{\textbf{BLIP}} & \multicolumn{1}{c|}{\textbf{VG}} & \multicolumn{1}{c|}{\textbf{OFA}} & \multicolumn{1}{c|}{\textbf{VT}} \\ \hline
\hline

\multirow{3}{*}{\begin{tabular}[c]{@{}c@{}}\textbf{MVSA} \\ (Acc)\end{tabular}} & \textbf{Flan-T5}   &  31.8    &  21.6  & 27.7  & 16.1  \\ \cline{2-6}
 & \textbf{T0pp}   &  62.3    &  61.8  & 62.4  & \textbf{63.1}  \\ \cline{2-6}
 & \textbf{OPT}   &  11.1    &  11.0  & 19.5  & 12.7  \\ \hline

\multirow{3}{*}{\begin{tabular}[c]{@{}c@{}}\textbf{OK-VQA} \\ (Acc)\end{tabular}} & \textbf{Flan-T5}   &  \textbf{34.4}    &  32.6  & 31.1  & 29.2  \\ \cline{2-6}
 & \textbf{T0pp}   &  20.2    &  19.7  & 18.3  & 17.8  \\ \cline{2-6}
 & \textbf{OPT}   &  3.9    &  4.0  & 19.5  & 14.8  \\ \hline

\multirow{3}{*}{\begin{tabular}[c]{@{}c@{}}\textbf{NLVR2} \\ (Acc)\end{tabular}} & \textbf{Flan-T5}   &  31.7    &  25.6  & 25.5  & 23.4  \\ \cline{2-6}
 & \textbf{T0pp}   &  50.7    &  49.4  & 50.1  & \textbf{51.0}  \\ \cline{2-6}
 & \textbf{OPT}   &  10.8    &  19.3  & 29.6  & 3.1  \\ \hline

\multirow{3}{*}{\begin{tabular}[c]{@{}c@{}}\textbf{HF} \\ (Acc)\end{tabular}} & \textbf{Flan-T5}   &  56.6    &  54.8  & 54.6  & \textbf{56.8}  \\ \cline{2-6}
 & \textbf{T0pp}   &  49.0    &  49.2  & 49.6  & 48.8  \\ \cline{2-6}
 & \textbf{OPT}   &  38.2    &  38.4  & 43.4  & 42.0  \\ \hline

\multirow{3}{*}{\begin{tabular}[c]{@{}c@{}}\textbf{MAMI} \\ (F1)\end{tabular}} & \textbf{Flan-T5}   &  \textbf{64.4}    &  48.6  & 60.2  & 60.3  \\ \cline{2-6}
 & \textbf{T0pp}   &  33.8    &  33.6  & 33.6  & 22.6  \\ \cline{2-6}
 & \textbf{OPT}   &  22.0    &  22.1  & 22.2  & 22.2  \\ \hline

\end{tabular}
\caption{Ablation study on the affect of using different image-as-text representation methods. All runs of each model has been set to \textit{adaptive} sample selection with \textit{n}=2 (number of in-context samples in a prompt). VG: ViT-GPT2, VT: Visual Tags}
\label{tab:captioning_results}
\end{table}

\subsection{Impact of In-Context Sample Selection}\label{subsec:impact_sample_selection}

\textbf{Sample Selection}: we have conducted experiments with different configurations of selecting in-context samples. The results are presented in Table~\ref{tab:in_context_samples}. In four out of five datasets, the \textit{adaptive} sample selection yields better performance than \textit{random} method. Only in the MVSA dataset, \textit{random} method yields the best result.

\textbf{Number of Samples}: The presented results for each evaluated language model include the different number of samples in a prompt. We tested numbers 0, 1, 2, 3 where \textit{n=0} essentially means that there are no in-context samples in a prompt, and it is \textbf{zero-shot performance} of an evaluated language model. It is a few-shot setting in cases where \textit{n} is bigger than zero. We can observe that in three out of five datasets, using $n>1$ yields better performance, whereas \textit{T0pp} achieves the best performance in \textit{MVSA} and \textit{NLVR2} datasets.

\subsection{Evaluation of Image-as-Text Representation Methods}

As explained in Section~\ref{subsec:image_context_methods}, we have used four methods of verbalising the visual content and adding the output to the prompt as an image-as-text representation. We have tested these methods on all datasets. The results are presented in Table~\ref{tab:captioning_results}. Based on the outcomes in Section~\ref{subsec:impact_sample_selection}, we have used \textit{adaptive} sample selection with \textit{n=2} for all runs. We can observe that in the majority of the evaluated datasets, using captions generated by \textit{BLIP} model yields the higher performance on average. The textual descriptions generated by the method \textit{Visual Tags} (collection of multiple image classification high-probability outputs) resulted in the highest performance on three datasets.

\begin{table}[!ht]
\centering
\small
\begin{tabular}{|c|c|c|}
\hline
\textbf{Dataset}     & \textbf{Models}           & \textbf{Result} \\ \hline

  \multirow{5}{*}{\begin{tabular}[c]{@{}c@{}}\textbf{MVSA} \\ (Acc)\end{tabular}} 
& Flan-T5, n=3, adaptive    & 35.2  \\ \cline{2-3} 
 & GPT-3, n=2, adaptive    & 64.3      \\ \cline{2-3} 
 & OPT, n=3, random      & 14.4     \\ \cline{2-3} 
  & T0pp, n=0, adaptive    & \textbf{68.1}       \\ \cline{2-3} 
  & \begin{tabular}[c]{@{}c@{}}\textbf{Fine-tuned V\&L Model}\\ \citet{DBLP:conf/mir/CheemaHME21}\end{tabular} & \textbf{\textit{75.3}}               \\ \hline

    \multirow{5}{*}{\begin{tabular}[c]{@{}c@{}}\textbf{OK-VQA} \\ (Acc)\end{tabular}} 
& Flan-T5, n=3, adaptive    & \textbf{35.4}  \\ \cline{2-3} 
 & GPT-3, n=2, adaptive    & 25.9      \\ \cline{2-3} 
 & OPT, n=1, adaptive      & 10.0     \\ \cline{2-3} 
  & T0pp, n=3, adaptive    &  20.3      \\ \cline{2-3} 
  & \begin{tabular}[c]{@{}c@{}}\textbf{Fine-tuned V\&L Model}\\ \citet{DBLP:conf/aaai/WuLSM22}\end{tabular} & \textbf{\textit{41.4} }              \\ \hline

  \multirow{5}{*}{\begin{tabular}[c]{@{}c@{}}\textbf{NLVR2} \\ (Acc)\end{tabular}} 
& Flan-T5, n=2, adaptive    & 31.7  \\ \cline{2-3} 
 & GPT-3, n=2, adaptive    & \textbf{63.0}      \\ \cline{2-3} 
 & OPT, n=1, random      & 47.7     \\ \cline{2-3} 
  & T0pp, n=1, adaptive    &  58.6      \\ \cline{2-3} 
  & \begin{tabular}[c]{@{}c@{}}\textbf{Fine-tuned V\&L Model}\\ \citet{DBLP:conf/eccv/ChenLYK0G0020}\end{tabular} & \textbf{\textit{79.5}}               \\ \hline

    \multirow{5}{*}{\begin{tabular}[c]{@{}c@{}}\textbf{HF} \\ (Acc)\end{tabular}} 
& Flan-T5, n=2, adaptive    & 57.4  \\ \cline{2-3} 
 & GPT-3, n=2, adaptive    & \textbf{58.8}      \\ \cline{2-3} 
 & OPT, n=1, random      & 44.4     \\ \cline{2-3} 
  & T0pp, n=1, adaptive    & 53.6      \\ \cline{2-3} 
  & \begin{tabular}[c]{@{}c@{}}\textbf{Fine-tuned V\&L Model}\\ \citet{DBLP:conf/nips/KielaFMGSRT20}\end{tabular} & \textbf{\textit{66.1}}               \\ \hline

\multirow{5}{*}{\begin{tabular}[c]{@{}c@{}}\textbf{MAMI} \\ (F1)\end{tabular}} 
& Flan-T5, n=2, adaptive    & 64.4  \\ \cline{2-3} 
 & GPT-3, n=2, adaptive    & \textbf{69.2}      \\ \cline{2-3} 
 & OPT, n=3, adaptive      & 22.4     \\ \cline{2-3} 
  & T0pp, n=2, adaptive    & 46.0       \\ \cline{2-3} 

  & \begin{tabular}[c]{@{}c@{}}\textbf{Fine-tuned V\&L Model}\\ \citet{zhang-wang-2022-srcb}\end{tabular} & \textbf{\textit{83.4 }}              \\ \hline

\end{tabular}
\caption{Overall comparison of the best-ranking configurations for each model. The best result for each dataset using prompting with language models is highlighted in bold. All model configurations use image captions generated by BLIP model. V\&L: Vision-Language}
\label{tab:overall_comparison}
\vspace{-0.5cm}
\end{table}

\subsection{Comparison of Language Models}

In Table~\ref{tab:overall_comparison}, we have selected the best-ranking configuration of each model for all datasets. All model configurations use image captions generated by the BLIP model to represent the image context in text. To reduce the budget, we ran GPT-3 experiments only on a pre-selected set of parameters (n=2, adaptive) that yielded the best results using open-source language models. 
The overall comparison of all results shows that GPT-3 achieves the best result on three datasets: \textit{MAMI}, \textit{HF}, and \textit{NLVR2}. T0pp achieves the best result on \textit{MVSA} dataset, whereas the best-ranking model for the \textit{OK-VQA} is Flan-T5. 

We have also included the results from the fine-tuned vision-language models for each dataset. By comparing the results obtained via prompting with fine-tuned models, with only a few-shots (\textit{n} = {1, 2, 3}), the language models can generalise to vision-language tasks and achieve comparable results. An important observation is that these models were trained only on text documents. Prompting these models on five downstream vision-language tasks by converting the visual content into textual representation made it possible.

\begin{figure*}[!ht]
  \includegraphics[width=\textwidth]{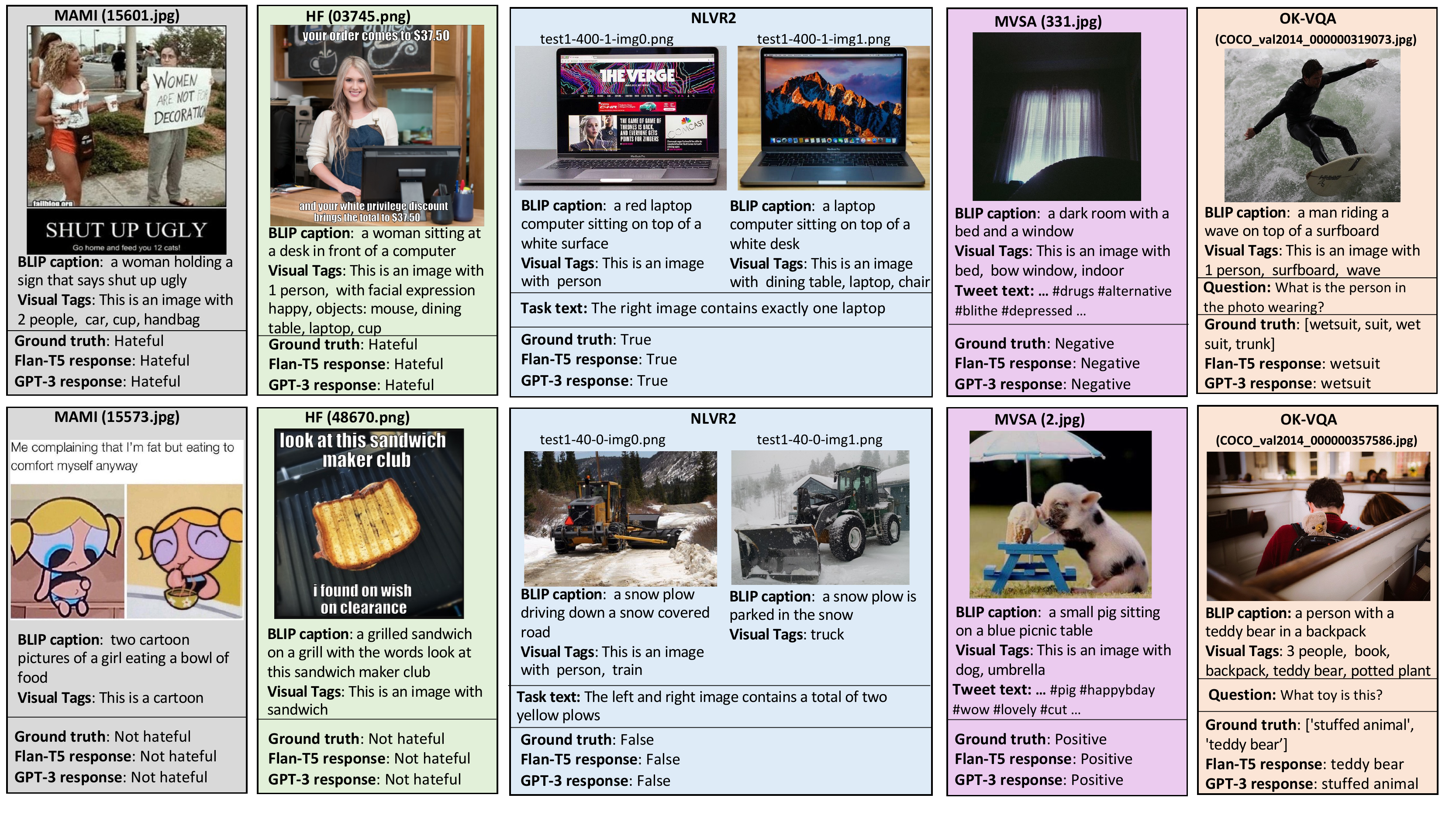}  \caption[width=\textwidth]{Qualitative examples for each evaluated dataset. The samples include the ground truth and responses generated via prompting Flan-T5 and GPT-3 models. Samples for the MAMI and HF datasets are prompted including the overlay text embedded in an image, which is excluded in the graphic for spacing reasons.}
  \label{fig:qualitative_examples}
\end{figure*}

\subsection{Qualitative Analysis}

We present qualitative examples from each dataset in Figure~\ref{fig:qualitative_examples}. Each sample includes the image-as-text representation extracted from the BLIP model. We also included the ground truth for each sample and the responses generated from Flan-T5 and GPT-3 models (best configurations as in Table~\ref{tab:overall_comparison}). We also added the Visual Tags for each sample (combinations of multiple image classification predictions) to show the the comparison against captions generated by the BLIP model. 

\subsection{Discussion}

We have presented experimental results that prompted large language models for five vision-language tasks. The prompting was made possible by representing the visual content in any task using methods such as image captioning or applying multiple image classification methods and combining their outputted high-ranking class labels. We have shown that such a method can achieve impressive results by presenting only two or three samples from a respective dataset compared with fine-tuned models on the entire train splits. It is worth mentioning that the gap between prompted models and fine-tuned models (in some evaluated datasets) is still there (margins of 10-20 points). One way of closing the such gap is by making the image-as-text representation methods achieve performance closer to how humans verbalise the visual content. Our paper essentially aims to highlight that given such image-as-text representations, which are only partial representations given the image models' capabilities, whether language models can be used for multimodal tasks by relying on their (imperfect) general reasoning mechanisms such as chain-of-thought~\citep{DBLP:journals/corr/abs-2201-11903}. Another way to achieve better performance (and close the gap with task-specific fine-tuned models) is to train vision-language models that are capable of in-context learning via prompting~\citep{DBLP:journals/corr/abs-2204-14198}.

We have also shown that the choice of in-context samples impacts the results. Using samples similar to the evaluated one (adaptive method) yields better performance than choosing them randomly.

Another critical observation to mention here is that different language models obtained the best results on various tasks. Overall, \textit{GPT-3} is the best-ranking model for three datasets. Among open-source models, \textit{T0pp} and \textit{Flan-T5} obtained the highest overall performance. Even though their performance was not the highest for many tasks, it is still possible to achieve comparable results or even the best ones in some cases. For the \textit{MVSA} dataset, \textit{T0pp} achieved the best performance even in a zero-shot setting. Thus, the language model's choice makes a difference in applying such models for any downstream tasks.

\section{Conclusion}\label{sec:conclusion}

In conclusion, our study has demonstrated the suitability and effectiveness of using large language models via prompting on vision-language tasks. Our approach relies on verbalising the visual content employing image captioning and classification models and prompts the language models along with the textual content. We have also shown that the choice of in-context samples and the method of verbalising the visual content impact the results. Our experimental evaluation suggests that this approach can achieve impressive results by presenting only a few samples from a dataset compared to models that are fine-tuned on entire train splits of the evaluated datasets. Furthermore, our study has also highlighted the importance of considering the choice of language models when applying them to such downstream tasks. We have demonstrated that different models perform better on various tasks, with GPT-3 achieving the highest overall performance across three datasets and open-source models T0pp and Flan-T5 achieving the best overall performance among them. Even though the performance of these models may not have been the best across all evaluated tasks, they still have the potential to be used in such cases and even achieve comparable results. For instance, T0pp yielded the best performance on the MVSA dataset, even in a zero-shot setting. Thus, the choice of language models is crucial for achieving optimal results in vision-language tasks.

\section*{Limitations}

\textbf{Limitations on the evaluated language models and obtained results}: The presented model architecture utilises various pre-trained language or image models. The main limitation of the experimental evaluation is not using other language models. Due to the limited budget and processing power, we have included the language models that have been shown to perform better based on the previous work. Another limitation is that we excluded language models that exceeded the 80 GB memory of an NVIDIA A100 GPU. Our experiments led to different results for the GPT-3 compared to \citet{DBLP:conf/aaai/YangGW0L0W22}. It can  be explained by using different methods for converting images to textual representations and slightly varying prompting structures.

\textbf{Limitations on the used image models}: The limitation concerning the pre-trained image models is that we selected a handful of methods based on their success for related tasks. Including other pre-trained models would increase the parameter space and thus increase the budget for the study.

\textbf{Limitations on the selected datasets}: All datasets are multimodal tasks where the underlying text is only in English. The choice of the dataset is related to the fact that there are limited multimodal datasets in other languages. The evaluation metric for the OK-VQA dataset requires the output to match exactly one of the expected answers. It counts as a wrong answer even if a slight change in the answer or another paraphrase is given as an output, e.g. ``race'' vs ``racing''. We applied the same evaluation criterion and left this improvement as future work.

\section*{Ethics Statement}
There might arise ethical issues as part of this work. The used pre-trained language models inherit particular biases as part of their learning process, which might affect the generated outputs. Another concern is the use of pre-trained image models for captioning or classification. The generated outputs from these models might predict certain visual concepts and thus leading to inaccurate text descriptions for the given images are generated. Another concern directly concerns using large language models as few-shot models. Such models have demonstrated high performance for many downstream tasks. However, the interpretation of the model predictions is still ongoing research.

\section*{Acknowledgements} We want to thank the anonymous reviewers for their comments. 
This work was supported by BMBF (German Federal Ministry of Research), project ``COCOBOTS'' (01IS21102A) and Deutsche Forschungsgemeinschaft (DFG, German Research Foundation) – 423217434 (``RECOLAGE'') grant.

\bibliography{references}

\begin{thebibliography}{49}
\expandafter\ifx\csname natexlab\endcsname\relax\def\natexlab#1{#1}\fi

\bibitem[{Alayrac et~al.(2022)Alayrac, Donahue, Luc, Miech, Barr, Hasson, Lenc,
  Mensch, Millican, Reynolds, Ring, Rutherford, Cabi, Han, Gong, Samangooei,
  Monteiro, Menick, Borgeaud, Brock, Nematzadeh, Sharifzadeh, Binkowski,
  Barreira, Vinyals, Zisserman, and
  Simonyan}]{DBLP:journals/corr/abs-2204-14198}
Jean{-}Baptiste Alayrac, Jeff Donahue, Pauline Luc, Antoine Miech, Iain Barr,
  Yana Hasson, Karel Lenc, Arthur Mensch, Katie Millican, Malcolm Reynolds,
  Roman Ring, Eliza Rutherford, Serkan Cabi, Tengda Han, Zhitao Gong, Sina
  Samangooei, Marianne Monteiro, Jacob Menick, Sebastian Borgeaud, Andrew
  Brock, Aida Nematzadeh, Sahand Sharifzadeh, Mikolaj Binkowski, Ricardo
  Barreira, Oriol Vinyals, Andrew Zisserman, and Karen Simonyan. 2022.
\newblock \href {https://doi.org/10.48550/arXiv.2204.14198} {Flamingo: a visual
  language model for few-shot learning}.
\newblock \emph{CoRR}, abs/2204.14198.

\bibitem[{Brown et~al.(2020)Brown, Mann, Ryder, and
  et~al.}]{DBLP:conf/nips/BrownMRSKDNSSAA20}
Tom~B. Brown, Benjamin Mann, Nick Ryder, and et~al. 2020.
\newblock \href
  {https://proceedings.neurips.cc/paper/2020/hash/1457c0d6bfcb4967418bfb8ac142f64a-Abstract.html}
  {Language models are few-shot learners}.
\newblock In \emph{Advances in Neural Information Processing Systems 33: Annual
  Conference on Neural Information Processing Systems 2020, NeurIPS 2020,
  December 6-12, 2020, virtual}.

\bibitem[{Carion et~al.(2020)Carion, Massa, Synnaeve, Usunier, Kirillov, and
  Zagoruyko}]{DBLP:conf/eccv/CarionMSUKZ20}
Nicolas Carion, Francisco Massa, Gabriel Synnaeve, Nicolas Usunier, Alexander
  Kirillov, and Sergey Zagoruyko. 2020.
\newblock \href {https://doi.org/10.1007/978-3-030-58452-8\_13} {End-to-end
  object detection with transformers}.
\newblock In \emph{Computer Vision - {ECCV} 2020 - 16th European Conference,
  Glasgow, UK, August 23-28, 2020, Proceedings, Part {I}}, volume 12346 of
  \emph{Lecture Notes in Computer Science}, pages 213--229. Springer.

\bibitem[{Cheema et~al.(2021)Cheema, Hakimov, M{\"{u}}ller{-}Budack, and
  Ewerth}]{DBLP:conf/mir/CheemaHME21}
Gullal~S. Cheema, Sherzod Hakimov, Eric M{\"{u}}ller{-}Budack, and Ralph
  Ewerth. 2021.
\newblock \href {https://doi.org/10.1145/3463945.3469058} {A fair and
  comprehensive comparison of multimodal tweet sentiment analysis methods}.
\newblock In \emph{MMPT@ICMR2021: Proceedings of the 2021 Workshop on
  Multi-Modal Pre-Training for Multimedia Understanding, Taipei, Taiwan, August
  21, 2021}, pages 37--45. {ACM}.

\bibitem[{Chen et~al.(2020)Chen, Li, Yu, Kholy, Ahmed, Gan, Cheng, and
  Liu}]{DBLP:conf/eccv/ChenLYK0G0020}
Yen{-}Chun Chen, Linjie Li, Licheng Yu, Ahmed~El Kholy, Faisal Ahmed, Zhe Gan,
  Yu~Cheng, and Jingjing Liu. 2020.
\newblock \href {https://doi.org/10.1007/978-3-030-58577-8\_7} {{UNITER:}
  universal image-text representation learning}.
\newblock In \emph{Computer Vision - {ECCV} 2020 - 16th European Conference,
  Glasgow, UK, August 23-28, 2020, Proceedings, Part {XXX}}, volume 12375 of
  \emph{Lecture Notes in Computer Science}, pages 104--120. Springer.

\bibitem[{Chung et~al.(2022)Chung, Hou, Longpre, Zoph, and
  et~al.}]{google-flan}
Hyung~Won Chung, Le~Hou, Shayne Longpre, Barret Zoph, and et~al. 2022.
\newblock \href {https://doi.org/10.48550/arXiv.2210.11416} {Scaling
  instruction-finetuned language models}.
\newblock \emph{CoRR}, abs/2210.11416.

\bibitem[{Devlin et~al.(2019)Devlin, Chang, Lee, and
  Toutanova}]{DBLP:conf/naacl/DevlinCLT19}
Jacob Devlin, Ming{-}Wei Chang, Kenton Lee, and Kristina Toutanova. 2019.
\newblock \href {https://doi.org/10.18653/v1/n19-1423} {{BERT:} pre-training of
  deep bidirectional transformers for language understanding}.
\newblock In \emph{Proceedings of the 2019 Conference of the North American
  Chapter of the Association for Computational Linguistics: Human Language
  Technologies, {NAACL-HLT} 2019, Minneapolis, MN, USA, June 2-7, 2019, Volume
  1 (Long and Short Papers)}, pages 4171--4186. Association for Computational
  Linguistics.

\bibitem[{Dong et~al.(2023)Dong, Li, Dai, Zheng, Wu, Chang, Sun, Xu, Li, and
  Sui}]{dong2022survey}
Qingxiu Dong, Lei Li, Damai Dai, Ce~Zheng, Zhiyong Wu, Baobao Chang, Xu~Sun,
  Jingjing Xu, Lei Li, and Zhifang Sui. 2023.
\newblock \href {https://doi.org/10.48550/arXiv.2301.00234} {A survey for
  in-context learning}.
\newblock \emph{CoRR}, abs/2301.00234.

\bibitem[{Dosovitskiy et~al.(2021)Dosovitskiy, Beyer, Kolesnikov, Weissenborn,
  Zhai, Unterthiner, Dehghani, Minderer, Heigold, Gelly, Uszkoreit, and
  Houlsby}]{DBLP:conf/iclr/DosovitskiyB0WZ21}
Alexey Dosovitskiy, Lucas Beyer, Alexander Kolesnikov, Dirk Weissenborn,
  Xiaohua Zhai, Thomas Unterthiner, Mostafa Dehghani, Matthias Minderer, Georg
  Heigold, Sylvain Gelly, Jakob Uszkoreit, and Neil Houlsby. 2021.
\newblock \href {https://openreview.net/forum?id=YicbFdNTTy} {An image is worth
  16x16 words: Transformers for image recognition at scale}.
\newblock In \emph{9th International Conference on Learning Representations,
  {ICLR} 2021, Virtual Event, Austria, May 3-7, 2021}. OpenReview.net.

\bibitem[{Fersini et~al.(2022)Fersini, Gasparini, Rizzi, Saibene, Chulvi,
  Rosso, Lees, and Sorensen}]{mami}
Elisabetta Fersini, Francesca Gasparini, Giulia Rizzi, Aurora Saibene, Berta
  Chulvi, Paolo Rosso, Alyssa Lees, and Jeffrey Sorensen. 2022.
\newblock {SemEval-2022 Task 5}: Multimedia automatic misogyny identification.
\newblock In \emph{Proceedings of the 16th International Workshop on Semantic
  Evaluation (SemEval-2022)}. Association for Computational Linguistics.

\bibitem[{Goodfellow et~al.(2015)Goodfellow, Erhan, Carrier, Courville, Mirza,
  Hamner, Cukierski, Tang, Thaler, Lee et~al.}]{goodfellow2015challenges}
Ian~J Goodfellow, Dumitru Erhan, Pierre~Luc Carrier, Aaron Courville, Mehdi
  Mirza, Ben Hamner, Will Cukierski, Yichuan Tang, David Thaler, Dong-Hyun Lee,
  et~al. 2015.
\newblock Challenges in representation learning: A report on three machine
  learning contests.
\newblock \emph{Neural Networks}, 64:59--63.

\bibitem[{Gui et~al.(2022)Gui, Wang, Huang, Hauptmann, Bisk, and
  Gao}]{gui-etal-2022-kat}
Liangke Gui, Borui Wang, Qiuyuan Huang, Alexander Hauptmann, Yonatan Bisk, and
  Jianfeng Gao. 2022.
\newblock \href {https://doi.org/10.18653/v1/2022.naacl-main.70} {{KAT}: A
  knowledge augmented transformer for vision-and-language}.
\newblock In \emph{Proceedings of the 2022 Conference of the North American
  Chapter of the Association for Computational Linguistics: Human Language
  Technologies}, pages 956--968, Seattle, United States. Association for
  Computational Linguistics.

\bibitem[{Kiela et~al.(2020)Kiela, Firooz, Mohan, Goswami, Singh, Ringshia, and
  Testuggine}]{DBLP:conf/nips/KielaFMGSRT20}
Douwe Kiela, Hamed Firooz, Aravind Mohan, Vedanuj Goswami, Amanpreet Singh,
  Pratik Ringshia, and Davide Testuggine. 2020.
\newblock \href
  {https://proceedings.neurips.cc/paper/2020/hash/1b84c4cee2b8b3d823b30e2d604b1878-Abstract.html}
  {The hateful memes challenge: Detecting hate speech in multimodal memes}.
\newblock In \emph{Advances in Neural Information Processing Systems 33: Annual
  Conference on Neural Information Processing Systems 2020, NeurIPS 2020,
  December 6-12, 2020, virtual}.

\bibitem[{Li et~al.(2022{\natexlab{a}})Li, Li, Xiong, and Hoi}]{blip}
Junnan Li, Dongxu Li, Caiming Xiong, and Steven C.~H. Hoi. 2022{\natexlab{a}}.
\newblock \href {https://proceedings.mlr.press/v162/li22n.html} {{BLIP:}
  bootstrapping language-image pre-training for unified vision-language
  understanding and generation}.
\newblock In \emph{International Conference on Machine Learning, {ICML} 2022,
  17-23 July 2022, Baltimore, Maryland, {USA}}, volume 162 of \emph{Proceedings
  of Machine Learning Research}, pages 12888--12900. {PMLR}.

\bibitem[{Li et~al.(2019)Li, Yatskar, Yin, Hsieh, and
  Chang}]{DBLP:journals/corr/abs-1908-03557}
Liunian~Harold Li, Mark Yatskar, Da~Yin, Cho{-}Jui Hsieh, and Kai{-}Wei Chang.
  2019.
\newblock \href {http://arxiv.org/abs/1908.03557} {Visualbert: {A} simple and
  performant baseline for vision and language}.
\newblock \emph{CoRR}, abs/1908.03557.

\bibitem[{Li et~al.(2022{\natexlab{b}})Li, Li, Shang, Dong, Sun, Liu, Ji,
  Jiang, and Liu}]{li-etal-2022-pre}
Shaobo Li, Xiaoguang Li, Lifeng Shang, Zhenhua Dong, Chengjie Sun, Bingquan
  Liu, Zhenzhou Ji, Xin Jiang, and Qun Liu. 2022{\natexlab{b}}.
\newblock \href {https://doi.org/10.18653/v1/2022.findings-acl.136} {How
  pre-trained language models capture factual knowledge? a causal-inspired
  analysis}.
\newblock In \emph{Findings of the Association for Computational Linguistics:
  ACL 2022}, pages 1720--1732, Dublin, Ireland. Association for Computational
  Linguistics.

\bibitem[{Liang et~al.(2022)Liang, Bommasani, Lee, Tsipras, and
  et~al.}]{DBLP:journals/corr/abs-2211-09110}
Percy Liang, Rishi Bommasani, Tony Lee, Dimitris Tsipras, and et~al. 2022.
\newblock \href {https://doi.org/10.48550/arXiv.2211.09110} {Holistic
  evaluation of language models}.
\newblock \emph{CoRR}, abs/2211.09110.

\bibitem[{Marino et~al.(2019)Marino, Rastegari, Farhadi, and
  Mottaghi}]{DBLP:conf/cvpr/MarinoRFM19}
Kenneth Marino, Mohammad Rastegari, Ali Farhadi, and Roozbeh Mottaghi. 2019.
\newblock \href {https://doi.org/10.1109/CVPR.2019.00331} {{OK-VQA:} {A} visual
  question answering benchmark requiring external knowledge}.
\newblock In \emph{{IEEE} Conference on Computer Vision and Pattern
  Recognition, {CVPR} 2019, Long Beach, CA, USA, June 16-20, 2019}, pages
  3195--3204. Computer Vision Foundation / {IEEE}.

\bibitem[{Merullo et~al.(2023)Merullo, Castricato, Eickhoff, and
  Pavlick}]{merullo2023linearly}
Jack Merullo, Louis Castricato, Carsten Eickhoff, and Ellie Pavlick. 2023.
\newblock \href {https://openreview.net/forum?id=8tYRqb05pVn} {Linearly mapping
  from image to text space}.
\newblock In \emph{The Eleventh International Conference on Learning
  Representations}.

\bibitem[{Min et~al.(2022)Min, Lyu, Holtzman, Artetxe, Lewis, Hajishirzi, and
  Zettlemoyer}]{DBLP:journals/corr/abs-2202-12837}
Sewon Min, Xinxi Lyu, Ari Holtzman, Mikel Artetxe, Mike Lewis, Hannaneh
  Hajishirzi, and Luke Zettlemoyer. 2022.
\newblock \href {http://arxiv.org/abs/2202.12837} {Rethinking the role of
  demonstrations: What makes in-context learning work?}
\newblock \emph{CoRR}, abs/2202.12837.

\bibitem[{Niu et~al.(2016)Niu, Zhu, Pang, and
  El{-}Saddik}]{DBLP:conf/mmm/NiuZPE16}
Teng Niu, Shiai Zhu, Lei Pang, and Abdulmotaleb El{-}Saddik. 2016.
\newblock \href {https://doi.org/10.1007/978-3-319-27674-8\_2} {Sentiment
  analysis on multi-view social data}.
\newblock In \emph{MultiMedia Modeling - 22nd International Conference, {MMM}
  2016, Miami, FL, USA, January 4-6, 2016, Proceedings, Part {II}}, volume 9517
  of \emph{Lecture Notes in Computer Science}, pages 15--27. Springer.

\bibitem[{{NLP Connect}(2022)}]{nlp_connect_2022}
{NLP Connect}. 2022.
\newblock \href {https://doi.org/10.57967/hf/0222} {vit-gpt2-image-captioning
  (revision 0e334c7)}.

\bibitem[{Quattoni and Torralba(2009)}]{DBLP:conf/cvpr/QuattoniT09}
Ariadna Quattoni and Antonio Torralba. 2009.
\newblock \href {https://doi.org/10.1109/CVPR.2009.5206537} {Recognizing indoor
  scenes}.
\newblock In \emph{2009 {IEEE} Computer Society Conference on Computer Vision
  and Pattern Recognition {(CVPR} 2009), 20-25 June 2009, Miami, Florida,
  {USA}}, pages 413--420. {IEEE} Computer Society.

\bibitem[{Radford et~al.(2021)Radford, Kim, Hallacy, Ramesh, Goh, Agarwal,
  Sastry, Askell, Mishkin, Clark, Krueger, and
  Sutskever}]{DBLP:conf/icml/RadfordKHRGASAM21}
Alec Radford, Jong~Wook Kim, Chris Hallacy, Aditya Ramesh, Gabriel Goh,
  Sandhini Agarwal, Girish Sastry, Amanda Askell, Pamela Mishkin, Jack Clark,
  Gretchen Krueger, and Ilya Sutskever. 2021.
\newblock \href {http://proceedings.mlr.press/v139/radford21a.html} {Learning
  transferable visual models from natural language supervision}.
\newblock In \emph{Proceedings of the 38th International Conference on Machine
  Learning, {ICML} 2021, 18-24 July 2021, Virtual Event}, volume 139 of
  \emph{Proceedings of Machine Learning Research}, pages 8748--8763. {PMLR}.

\bibitem[{Radford et~al.(2018)Radford, Narasimhan, Salimans, and
  Sutskever}]{gpt}
Alec Radford, Karthik Narasimhan, Tim Salimans, and Ilya Sutskever. 2018.
\newblock \href {https://openai.com/blog/language-unsupervised/} {Improving
  language understanding by generative pre-training}.

\bibitem[{Radford et~al.(2019)Radford, Wu, Child, Luan, Amodei, Sutskever, and
  et~al.}]{gpt2}
Alec Radford, Jeffrey Wu, Rewon Child, David Luan, Dario Amodei, Ilya
  Sutskever, and et~al. 2019.
\newblock Language models are unsupervised multitask learners.

\bibitem[{Reimers and Gurevych(2019)}]{DBLP:conf/emnlp/ReimersG19}
Nils Reimers and Iryna Gurevych. 2019.
\newblock \href {https://doi.org/10.18653/v1/D19-1410} {Sentence-bert: Sentence
  embeddings using siamese bert-networks}.
\newblock In \emph{Proceedings of the 2019 Conference on Empirical Methods in
  Natural Language Processing and the 9th International Joint Conference on
  Natural Language Processing, {EMNLP-IJCNLP} 2019, Hong Kong, China, November
  3-7, 2019}, pages 3980--3990. Association for Computational Linguistics.

\bibitem[{Sanh et~al.(2022)Sanh, Webson, Raffel, Bach, Sutawika, and
  et~al.}]{DBLP:conf/iclr/SanhWRBSACSRDBX22}
Victor Sanh, Albert Webson, Colin Raffel, Stephen Bach, Lintang Sutawika, and
  et~al. 2022.
\newblock \href {https://openreview.net/forum?id=9Vrb9D0WI4} {Multitask
  prompted training enables zero-shot task generalization}.
\newblock In \emph{The Tenth International Conference on Learning
  Representations, {ICLR} 2022, Virtual Event, April 25-29, 2022}.
  OpenReview.net.

\bibitem[{Scao et~al.(2022)Scao, Fan, Akiki, Pavlick, Ilic, and et~al.}]{bloom}
Teven~Le Scao, Angela Fan, Christopher Akiki, Ellie Pavlick, Suzana Ilic, and
  et~al. 2022.
\newblock \href {https://doi.org/10.48550/arXiv.2211.05100} {{BLOOM:} {A}
  176b-parameter open-access multilingual language model}.
\newblock \emph{CoRR}, abs/2211.05100.

\bibitem[{Suhr et~al.(2019)Suhr, Zhou, Zhang, Zhang, Bai, and
  Artzi}]{DBLP:conf/acl/SuhrZZZBA19}
Alane Suhr, Stephanie Zhou, Ally Zhang, Iris Zhang, Huajun Bai, and Yoav Artzi.
  2019.
\newblock \href {https://doi.org/10.18653/v1/p19-1644} {A corpus for reasoning
  about natural language grounded in photographs}.
\newblock In \emph{Proceedings of the 57th Conference of the Association for
  Computational Linguistics, {ACL} 2019, Florence, Italy, July 28- August 2,
  2019, Volume 1: Long Papers}, pages 6418--6428. Association for Computational
  Linguistics.

\bibitem[{Talmor et~al.(2020)Talmor, Elazar, Goldberg, and
  Berant}]{DBLP:journals/tacl/TalmorEGB20}
Alon Talmor, Yanai Elazar, Yoav Goldberg, and Jonathan Berant. 2020.
\newblock \href {https://doi.org/10.1162/tacl\_a\_00342} {olmpics - on what
  language model pre-training captures}.
\newblock \emph{Trans. Assoc. Comput. Linguistics}, 8:743--758.

\bibitem[{Tsimpoukelli et~al.(2021)Tsimpoukelli, Menick, Cabi, Eslami, Vinyals,
  and Hill}]{DBLP:conf/nips/TsimpoukelliMCE21}
Maria Tsimpoukelli, Jacob Menick, Serkan Cabi, S.~M.~Ali Eslami, Oriol Vinyals,
  and Felix Hill. 2021.
\newblock \href
  {https://proceedings.neurips.cc/paper/2021/hash/01b7575c38dac42f3cfb7d500438b875-Abstract.html}
  {Multimodal few-shot learning with frozen language models}.
\newblock In \emph{Advances in Neural Information Processing Systems 34: Annual
  Conference on Neural Information Processing Systems 2021, NeurIPS 2021,
  December 6-14, 2021, virtual}, pages 200--212.

\bibitem[{Vaswani et~al.(2017)Vaswani, Shazeer, Parmar, Uszkoreit, Jones,
  Gomez, Kaiser, and Polosukhin}]{DBLP:conf/nips/VaswaniSPUJGKP17}
Ashish Vaswani, Noam Shazeer, Niki Parmar, Jakob Uszkoreit, Llion Jones,
  Aidan~N. Gomez, Lukasz Kaiser, and Illia Polosukhin. 2017.
\newblock \href
  {https://proceedings.neurips.cc/paper/2017/hash/3f5ee243547dee91fbd053c1c4a845aa-Abstract.html}
  {Attention is all you need}.
\newblock In \emph{Advances in Neural Information Processing Systems 30: Annual
  Conference on Neural Information Processing Systems 2017, December 4-9, 2017,
  Long Beach, CA, {USA}}, pages 5998--6008.

\bibitem[{Vrandecic and Kr{\"{o}}tzsch(2014)}]{DBLP:journals/cacm/VrandecicK14}
Denny Vrandecic and Markus Kr{\"{o}}tzsch. 2014.
\newblock \href {https://doi.org/10.1145/2629489} {Wikidata: a free
  collaborative knowledgebase}.
\newblock \emph{Commun. {ACM}}, 57(10):78--85.

\bibitem[{Wang and Komatsuzaki(2021)}]{gpt-j}
Ben Wang and Aran Komatsuzaki. 2021.
\newblock {GPT-J-6B: A 6 Billion Parameter Autoregressive Language Model}.
\newblock \url{https://github.com/kingoflolz/mesh-transformer-jax}.

\bibitem[{Wang et~al.(2022{\natexlab{a}})Wang, Yang, Men, Lin, Bai, Li, Ma,
  Zhou, Zhou, and Yang}]{DBLP:conf/icml/WangYMLBLMZZY22}
Peng Wang, An~Yang, Rui Men, Junyang Lin, Shuai Bai, Zhikang Li, Jianxin Ma,
  Chang Zhou, Jingren Zhou, and Hongxia Yang. 2022{\natexlab{a}}.
\newblock \href {https://proceedings.mlr.press/v162/wang22al.html} {{OFA:}
  unifying architectures, tasks, and modalities through a simple
  sequence-to-sequence learning framework}.
\newblock In \emph{International Conference on Machine Learning, {ICML} 2022,
  17-23 July 2022, Baltimore, Maryland, {USA}}, volume 162 of \emph{Proceedings
  of Machine Learning Research}, pages 23318--23340. {PMLR}.

\bibitem[{Wang et~al.(2022{\natexlab{b}})Wang, Li, Xu, Zhou, Lei, Lin, Wang,
  Yang, Zhu, Hoiem, Chang, Bansal, and Ji}]{DBLP:journals/corr/abs-2205-10747}
Zhenhailong Wang, Manling Li, Ruochen Xu, Luowei Zhou, Jie Lei, Xudong Lin,
  Shuohang Wang, Ziyi Yang, Chenguang Zhu, Derek Hoiem, Shih{-}Fu Chang, Mohit
  Bansal, and Heng Ji. 2022{\natexlab{b}}.
\newblock Language models with image descriptors are strong few-shot
  video-language learners.
\newblock In \emph{Advances in Neural Information Processing Systems}.

\bibitem[{Wei et~al.(2022{\natexlab{a}})Wei, Bosma, Zhao, Guu, Yu, Lester, Du,
  Dai, and Le}]{DBLP:conf/iclr/WeiBZGYLDDL22}
Jason Wei, Maarten Bosma, Vincent~Y. Zhao, Kelvin Guu, Adams~Wei Yu, Brian
  Lester, Nan Du, Andrew~M. Dai, and Quoc~V. Le. 2022{\natexlab{a}}.
\newblock \href {https://openreview.net/forum?id=gEZrGCozdqR} {Finetuned
  language models are zero-shot learners}.
\newblock In \emph{The Tenth International Conference on Learning
  Representations, {ICLR} 2022, Virtual Event, April 25-29, 2022}.
  OpenReview.net.

\bibitem[{Wei et~al.(2022{\natexlab{b}})Wei, Wang, Schuurmans, Bosma, Chi, Le,
  and Zhou}]{DBLP:journals/corr/abs-2201-11903}
Jason Wei, Xuezhi Wang, Dale Schuurmans, Maarten Bosma, Ed~H. Chi, Quoc Le, and
  Denny Zhou. 2022{\natexlab{b}}.
\newblock Chain of thought prompting elicits reasoning in large language
  models.
\newblock In \emph{Advances in Neural Information Processing Systems}.

\bibitem[{Wu et~al.(2020)Wu, Xu, Dai, Wan, Zhang, Tomizuka, Keutzer, and
  Vajda}]{DBLP:journals/corr/abs-2006-03677}
Bichen Wu, Chenfeng Xu, Xiaoliang Dai, Alvin Wan, Peizhao Zhang, Masayoshi
  Tomizuka, Kurt Keutzer, and Peter Vajda. 2020.
\newblock \href {http://arxiv.org/abs/2006.03677} {Visual transformers:
  Token-based image representation and processing for computer vision}.
\newblock \emph{CoRR}, abs/2006.03677.

\bibitem[{Wu et~al.(2022)Wu, Lu, Sabharwal, and
  Mottaghi}]{DBLP:conf/aaai/WuLSM22}
Jialin Wu, Jiasen Lu, Ashish Sabharwal, and Roozbeh Mottaghi. 2022.
\newblock \href {https://ojs.aaai.org/index.php/AAAI/article/view/20174}
  {Multi-modal answer validation for knowledge-based {VQA}}.
\newblock In \emph{Thirty-Sixth {AAAI} Conference on Artificial Intelligence,
  {AAAI} 2022, Thirty-Fourth Conference on Innovative Applications of
  Artificial Intelligence, {IAAI} 2022, The Twelveth Symposium on Educational
  Advances in Artificial Intelligence, {EAAI} 2022 Virtual Event, February 22 -
  March 1, 2022}, pages 2712--2721. {AAAI} Press.

\bibitem[{Yang et~al.(2022)Yang, Gan, Wang, Hu, Lu, Liu, and
  Wang}]{DBLP:conf/aaai/YangGW0L0W22}
Zhengyuan Yang, Zhe Gan, Jianfeng Wang, Xiaowei Hu, Yumao Lu, Zicheng Liu, and
  Lijuan Wang. 2022.
\newblock \href {https://ojs.aaai.org/index.php/AAAI/article/view/20215} {An
  empirical study of {GPT-3} for few-shot knowledge-based {VQA}}.
\newblock In \emph{Thirty-Sixth {AAAI} Conference on Artificial Intelligence,
  {AAAI} 2022, Thirty-Fourth Conference on Innovative Applications of
  Artificial Intelligence, {IAAI} 2022, The Twelveth Symposium on Educational
  Advances in Artificial Intelligence, {EAAI} 2022 Virtual Event, February 22 -
  March 1, 2022}, pages 3081--3089. {AAAI} Press.

\bibitem[{Zeng et~al.(2022)Zeng, Wong, Welker, Choromanski, Tombari, Purohit,
  Ryoo, Sindhwani, Lee, Vanhoucke, and
  Florence}]{DBLP:journals/corr/abs-2204-00598}
Andy Zeng, Adrian Wong, Stefan Welker, Krzysztof Choromanski, Federico Tombari,
  Aveek Purohit, Michael~S. Ryoo, Vikas Sindhwani, Johnny Lee, Vincent
  Vanhoucke, and Pete Florence. 2022.
\newblock \href {https://doi.org/10.48550/arXiv.2204.00598} {Socratic models:
  Composing zero-shot multimodal reasoning with language}.
\newblock \emph{CoRR}, abs/2204.00598.

\bibitem[{Zhang and Wang(2022)}]{zhang-wang-2022-srcb}
Jing Zhang and Yujin Wang. 2022.
\newblock \href {https://doi.org/10.18653/v1/2022.semeval-1.81} {{SRCB} at
  {S}em{E}val-2022 task 5: Pretraining based image to text late sequential
  fusion system for multimodal misogynous meme identification}.
\newblock In \emph{Proceedings of the 16th International Workshop on Semantic
  Evaluation (SemEval-2022)}, pages 585--596, Seattle, United States.
  Association for Computational Linguistics.

\bibitem[{Zhang et~al.(2016)Zhang, Zhang, Li, and Qiao}]{7553523}
Kaipeng Zhang, Zhanpeng Zhang, Zhifeng Li, and Yu~Qiao. 2016.
\newblock \href {https://doi.org/10.1109/LSP.2016.2603342} {Joint face
  detection and alignment using multitask cascaded convolutional networks}.
\newblock \emph{IEEE Signal Processing Letters}, 23(10):1499--1503.

\bibitem[{Zhang et~al.(2022)Zhang, Roller, Goyal, Artetxe, Chen, and
  et~al.}]{opt}
Susan Zhang, Stephen Roller, Naman Goyal, Mikel Artetxe, Moya Chen, and et~al.
  2022.
\newblock \href {http://arxiv.org/abs/2205.01068} {Opt: Open pre-trained
  transformer language models}.

\bibitem[{Zhou et~al.(2018)Zhou, Lapedriza, Khosla, Oliva, and
  Torralba}]{DBLP:journals/pami/ZhouLKO018}
Bolei Zhou, {\`{A}}gata Lapedriza, Aditya Khosla, Aude Oliva, and Antonio
  Torralba. 2018.
\newblock \href {https://doi.org/10.1109/TPAMI.2017.2723009} {Places: {A} 10
  million image database for scene recognition}.
\newblock \emph{{IEEE} Trans. Pattern Anal. Mach. Intell.}, 40(6):1452--1464.

\bibitem[{Zhou et~al.(2022{\natexlab{a}})Zhou, Yang, Loy, and
  Liu}]{DBLP:conf/cvpr/ZhouYL022}
Kaiyang Zhou, Jingkang Yang, Chen~Change Loy, and Ziwei Liu.
  2022{\natexlab{a}}.
\newblock \href {https://doi.org/10.1109/CVPR52688.2022.01631} {Conditional
  prompt learning for vision-language models}.
\newblock In \emph{{IEEE/CVF} Conference on Computer Vision and Pattern
  Recognition, {CVPR} 2022, New Orleans, LA, USA, June 18-24, 2022}, pages
  16795--16804. {IEEE}.

\bibitem[{Zhou et~al.(2022{\natexlab{b}})Zhou, Yang, Loy, and
  Liu}]{DBLP:journals/ijcv/ZhouYLL22}
Kaiyang Zhou, Jingkang Yang, Chen~Change Loy, and Ziwei Liu.
  2022{\natexlab{b}}.
\newblock \href {https://doi.org/10.1007/s11263-022-01653-1} {Learning to
  prompt for vision-language models}.
\newblock \emph{Int. J. Comput. Vis.}, 130(9):2337--2348.

\end{thebibliography}
\bibliographystyle{acl_natbib}

\appendix
\begin{figure*}[!ht]
 \centering
 \begin{subfigure}[b]{\textwidth}
     \centering
     \includegraphics[scale=0.53]{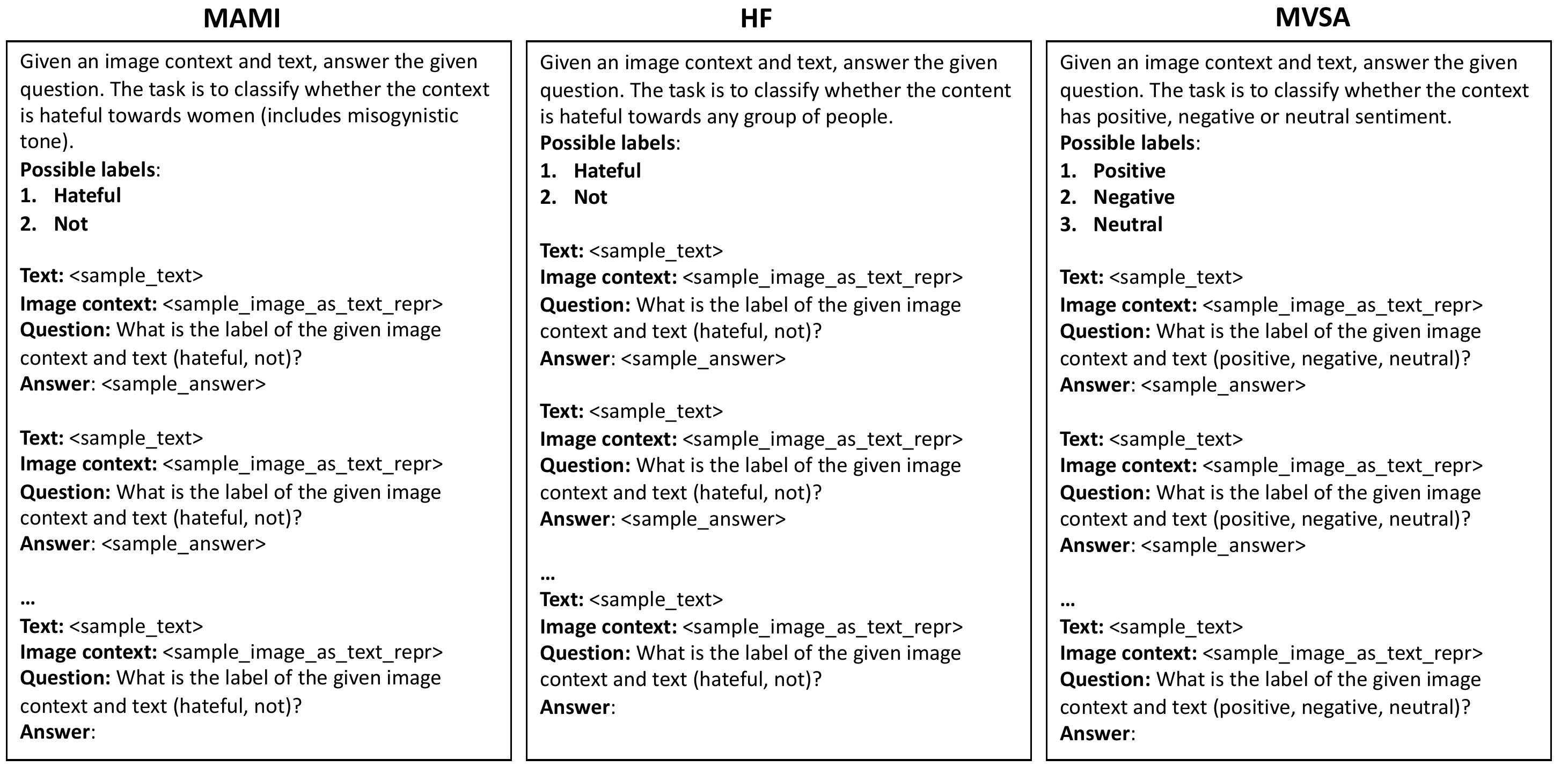}
     \caption{Prompt structures for MAMI, HF and MVSA datasets}
     \label{fig:prompt_structure1}
 \end{subfigure}
 \begin{subfigure}[b]{\textwidth}
     \centering
     \includegraphics[scale=0.53]{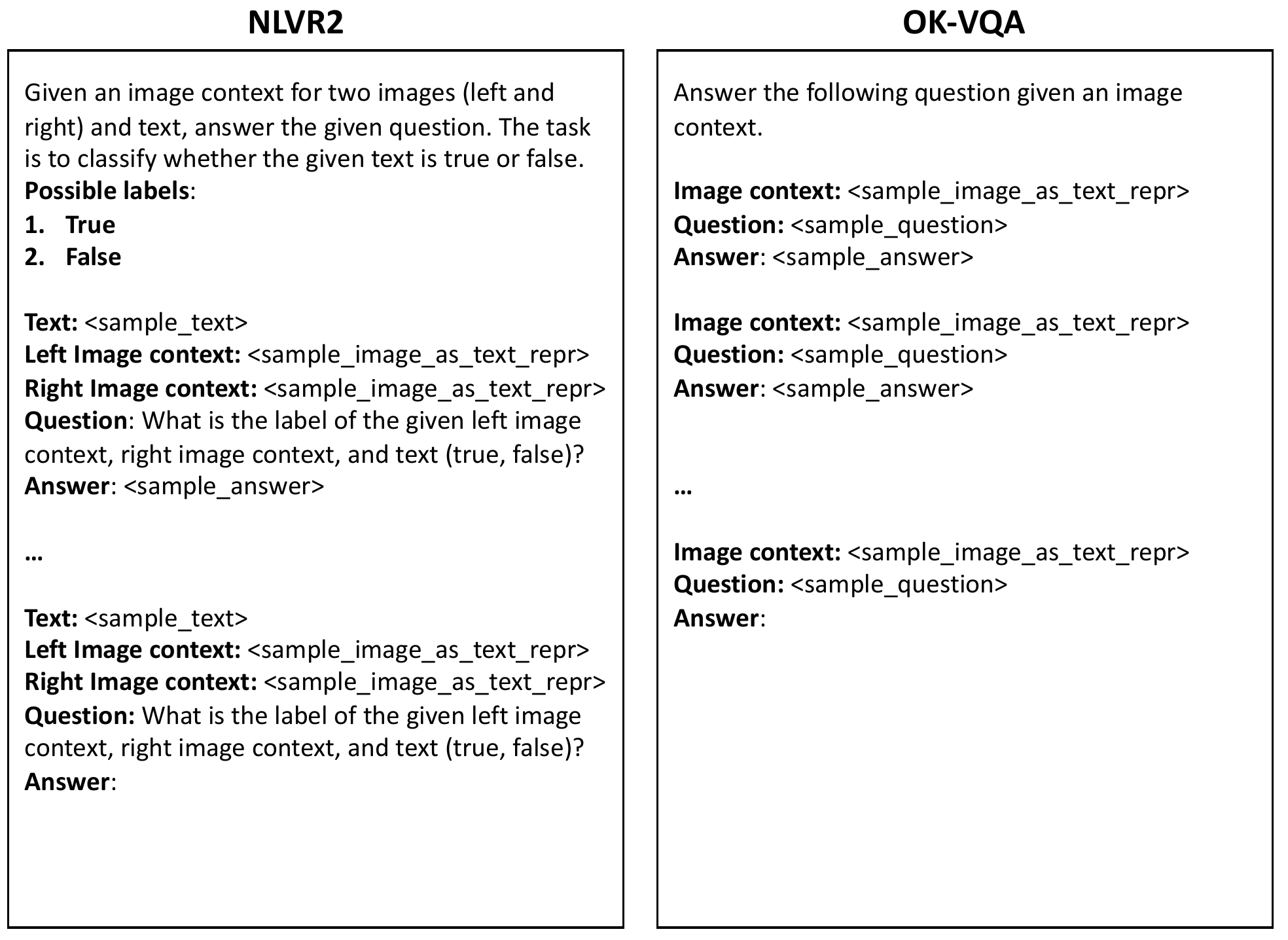}
     \caption{Prompt structures for NLVR2, OK-VQA datasets}
     \label{fig:prompt_structure2}
 \end{subfigure}
 \caption{Prompt structures for each evaluated dataset. Each prompt structure includes a task description, which also includes possible labels, selected \textit{n} in-context samples, and the evaluated sample. The prompted language models are expected to generated the next text sequence that starts after the last occurrence of the word \textit{Answer}.}
 \label{fig:prompt_structures}
\end{figure*}

\section{Prompt Structures}\label{sec:prompt_appendix}

We provided the prompt structures for all datasets in Figure~\ref{fig:prompt_structures}. Given a sample from a respective dataset, the prompt structures are initialised to create a prompt text. Each prompt text includes a task description followed by task-specific labels (only for classification tasks). In the middle of the prompt text are the selected in-context samples. The bottom part includes the evaluation sample, which is represented by only its input text, image-as-text representation (image context), and the task-specific question. The prompted language model is expected to generate the next text sequence that starts after the last occurrence of the word \textit{Answer}.

\end{document}